\documentclass[11pt,a4paper]{article}
\usepackage[hyperref]{eacl2021}
\usepackage{times}
\usepackage{latexsym}

\usepackage{graphicx}
\graphicspath{ {./Images/} }
\usepackage{booktabs}
\usepackage{footmisc}
\usepackage{float}
\usepackage{multirow}
\usepackage{enumitem}
\usepackage{caption}
\usepackage{makecell}
\usepackage{multicol}
\usepackage{amsmath}

\usepackage{microtype}

\aclfinalcopy 

\title{Improving Distantly-Supervised Relation Extraction through BERT-based Label \& Instance Embeddings}

\author{Despina Christou \\
  School of Informatics, \\
  Aristotle University of Thessaloniki, \\
  54124, Greece\\
  \texttt{christoud@csd.auth.gr} \\
  \And
  Grigorios Tsoumakas \\
  School of Informatics, \\
  Aristotle University of Thessaloniki,\\
  54124, Greece\\
  \texttt{greg@csd.auth.gr} \\}

\date{}

\begin{document}
\maketitle


\begin{abstract}
    Distantly-supervised relation extraction (RE) is an effective method to scale RE to large corpora but suffers from noisy labels. Existing approaches try to alleviate noise through multi-instance learning and by providing additional information, but manage to recognize mainly the top frequent relations, neglecting those in the long-tail. We propose REDSandT (Relation Extraction with Distant Supervision and Transformers), a novel distantly-supervised transformer-based RE method, that manages to capture a wider set of relations through highly informative instance and label embeddings for RE, by exploiting BERT's pre-trained model, and the relationship between labels and entities, respectively. We guide REDSandT to focus solely on relational tokens by fine-tuning BERT on a structured input, including the sub-tree connecting an entity pair and the entities' types. Using the extracted informative vectors, we shape label embeddings, which we also use as attention mechanism over instances to further reduce noise. Finally, we represent sentences by concatenating relation and instance embeddings. Experiments in the NYT-10 dataset show that REDSandT captures a broader set of relations with higher confidence, achieving state-of-the-art AUC (0.424). 
\end{abstract}

\section{Introduction}
\label{intro}
Relation Extraction (RE) aims to detect semantic relationships between entity pairs in natural texts and has proven to be crucial in various natural language processing (NLP) applications, including question answering, and knowledge-base (KB) population.

Most RE methods follow a supervised approach, with the required number of labeled training data rendering the whole process time and labor-intensive. To automatically construct datasets for RE, \cite{Mintz2009} proposed to use distant supervision (DS) from a KB, assuming that if two entities exhibit a relationship in a KB, then all sentences mentioning these entities express this relation. Inevitably, this assumption generates false-positives and leads distantly-created datasets to contain erroneous labels. To alleviate the \textit{wrong labeling problem}, \cite{Riedel2010} relaxed this assumption so that it does not hold for all instances and along with \cite{Hoffmann_2011, Surdeanu2012} proposed multi-instance based learning. Under this setting, classification shifts from instance-level to bag-level, with a bag consisting of all instances that contain a specific entity pair.

Current state-of-the-art RE methods try to reduce the effect of noisy instances by: i) identifying valid instances through multi-instance learning and selective attention \cite{Lin2016}, ii) reducing inner-sentence noise by capturing long-range dependencies using syntactic information from dependency parses \cite{Mintz2009, He2018, Liu2018}, specialized models like piecewise CNN (PCNN) and graph CNN (GCNN), or word-level attention \cite{He2018}, and iii) enhancing model effectiveness using external knowledge (i.e. KB entity types \cite{Vashishth2018}, entity descriptions \cite{Ji2017, Hu2019}, relation phrases \cite{Vashishth2018}) or transfer knowledge from pre-trained models \cite{Alt2019}.

The study of the above approaches led us to the following core observations. First, among all models used in the literature, the use of a pretrained transformer-based language model (LM) can help in recognizing a broader set of relations, even though at the expense of time and computational resources, and second, the relationship between label and entities can entail valuable information but rarely used over external knowledge. Driven by these observations we inspired to develop a novel transformer-based model that can efficiently capture instance and label embeddings in less complexity so as to drive RE in recognizing a broader set of relations.


We propose REDSandT (Relation Extraction with Distant Supervision and Transformers), a novel transformer-based RE model for distant supervision. To handle the problem of noisy instances, we guide REDSandT to focus solely on relational tokens by fine-tuning BERT on a structured input, including the sub-tree connecting an entity pair (STP) and the entities' types. The input's RE-specific formation, along with BERT's knowledge from unsupervised pre-training, results in REDSandT generating informative vectors. Using these vectors, we shape relation embeddings representing the entities' distance in vector space. Relation embeddings are then used as relation-wise attention over instance representation to reduce the effect of less-informative tokens. Finally, REDSandT encodes sentences by concatenating relation and weighted-instance embeddings, with relation classification to occur at bag-level as a weighted sum over its sentences' predictions. 

We chose BERT over other transformer-based models because it considers bidirectionality while training. We assume that this characteristic is important to efficiently capture entities' interactions without requiring an additional task that importantly increases complexity (i.e. fine-tuning an auxiliary objective in GPT \cite{Alt2019}).

The main contributions of this paper can be summarized as follows:
\begin{itemize}[noitemsep,nolistsep]
    \item We extend BERT to handle multi-instance learning to directly fine-tune the model in a DS setting and reduce error accumulation.
    \item Relation embeddings captured through BERT fine-tuned on our RE-specific input help to recognize a wider set of relations, including relations in the long-tail.
    \item Suppressing the input sentence to its relational tokens through STP encoding allowed us to capture informative instance embeddings while preserving low complexity to train our model on modest hardware.
    \item Experiments on the NYT-10 dataset show REDSandT to surpass state-of-the-art models \cite{Vashishth2018, Alt2019} in AUC (1.0 \& 0.2 units respectively) and performance at higher recall values, while achieving a 7-10\% improvement in P@\{100,200,300\} over \cite{Alt2019}.
\end{itemize}

\section{REDSandT}
\label{sect:redsandt}
Given a bag of sentences $\{s_1, s_2, ..., s_n\}$ that concern a specific entity pair, REDSandT generates a probability distribution on the set of possible relations. REDSandT utilizes BERT pre-trained LM to capture the semantic and syntactic features of sentences by transferring pre-trained common-sense knowledge. We extend BERT to handle multi-instance learning, and we fine-tune the model to classify the relation linking the entity pair given the associated sentences. 

During fine-tuning, we employ a structured, RE-specific input to minimize architectural changes to the model \cite{Radford2018}. Each sentence is adapted to a structured text, including the sentences' tokens connecting the entity pair (STP) along with the entities types. We transform the input into a (sub-)word-level distributed representation using BPE and positional embeddings from BERT fine-tuned on our corpora. Then, we form final sentence representation by concatenating relation embedding and sentence representation weighted with the relation embedding. Lastly, we use attention over the bag's sentences to shape bag representation, which is then fed to a softmax layer to get the bag 's relation distribution. 

REDSandT can be summarized in three components, namely sentence encoder, bag encoder, and model training. Each component is described in detail in the following sections with the overall architecture shown in Figure \ref{fig:redsandt_input} and \ref{fig:redsandt_architecture}.

\subsection{Sentence Encoder}
\begin{figure*}
    \centering
    \includegraphics[scale=0.47]{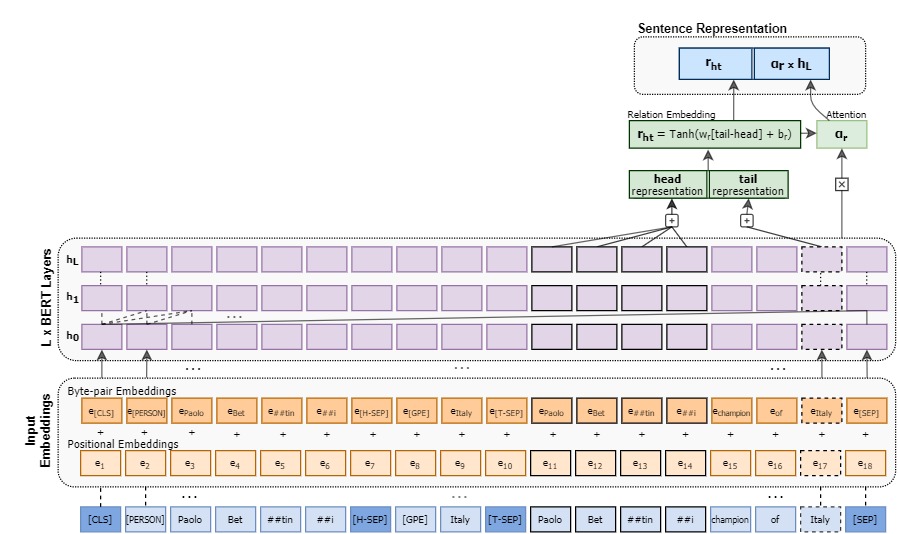}
    \caption{Sentence Representation in REDSandT. The input embedding $h_0$ to BERT is created by summing over the positional and byte pair embeddings for each token in the structured input. States $h_t$ are obtained by self-attending over the states of the previous layer $h_{t-1}$. Final sentence representation is obtained by concatenating the relation embedding $r_{ht}$, and the final fine-tuned BERT layer $h_L$ weighted with relation attention $\alpha_r$. Head and tail tokens participating in the relation embedding formation are marked with bold and dashed lines respectively.}
    \label{fig:redsandt_input}
    \vspace{-4mm}
\end{figure*}

Given a sentence $x$ and an entity pair $\langle h, t \rangle$, REDSandT constructs a distributed representation of the sentence by concatenating relation and instance embeddings. Overall sentence encoding is represented in Figure \ref{fig:redsandt_input}, with following sections to examine the sentence encoder parts in a bottom-up way.

\subsubsection{Input Representation}
\label{subsec:inputRepresentation}
Relation extraction requires a structured input that can sufficiently capture the latent relation between an entity pair and its surrounding text. Our input representation encodes each sentence as a sequence of tokens, depicted in the very bottom of Figure \ref{fig:redsandt_input}. 

It starts with the head entity type and token(s) followed by delimiter [H-SEP], continues with the tail entity type, and token(s) followed by delimiter [T-SEP] and ends with the token sequence of the sentence's STP path. The whole input starts and ends with special delimiters [CLS] and [SEP], respectively. In BERT, [CLS] typically acts as a pooling token representing the whole sequence for downstream tasks, such as RE. 

Several other sentence encodings were attempted\footnote{Trials included encoding overall sentence tokens, STP tokens only, SDP (\cite{Xu2015}) tokens only, using common $\langle h, t \rangle$ delimiter, using single delimiter between entities and STP, removing entity type information.} with the presented one to perform the best. Moreover, the ablation studies in section \ref{sec:ablation_studies}, reveal the importance of encoding entities' types and compressing the original sentence to the below-presented STP path.
Below, we present in brief how we form the sub-tree parse of the input and the entity types. 
\newline
\textbf{Sub-tree parse of input sentence}:
We utilize the sub-tree parse (STP) of the input sentence in order to reduce the noisy words within sentence and focus on the relational tokens. Precisely, STP preserves the path of the sentence that connects the two entities with their least common ancestor (LCA)'s parent. Compared to other implementations \cite{Liu2018}, who shape the final STP sequence by re-assigning the participating tokens into their original sequence order, we preserve the tokens' order within STP achieving a grammatical normalization of the original sentence.
\newline
\textbf{Entity Type special tokens}:
In the extent that every relation puts some constraint on the type of participating entities \cite{Liu2014,Vashishth2018}, we incorporate the entity type in the model's structured input (see bottom of Figure \ref{fig:redsandt_input}). 
\newline
Precisely, we incorporate 18 generic entity types, captured from recognizing NYT-10 sentence’s entities with the spaCy model\footnote{\href{https://spacy.io/models/en}{https://spacy.io/models/en}}. We assume these types KB-independent and easily accessible with our experiments in section \ref{sec:ablation_studies} indicating their inclusion to improve performance.

\subsubsection{Input Embeddings}
The input embedding $h_0$ to BERT is created by summing over the positional and byte pair embeddings for each token in the structured input. 
\newline
\textbf{Byte-pair tokens encoding}:
To make use of sub-word information, we tokenize input using byte-pair encoding (BPE) \cite{Sennrich2016}. We particularly use the tokenizer from the pre-trained model (30,000 tokens), which we extend with 20 task-specific tokens (e.g., [H-SEP], [T-SEP], and the 18 entity type tokens). Added tokens serve a special meaning in the input representation, thus are not split into sub-words by the tokenizer.
\newline
\textbf{Positional encoding}:
Positional encoding is an essential part of BERT's attention mechanism. Precisely, BERT learns a unique position embedding to represent each of the input (sub-word) token positions within the sequence.

\subsubsection{Sentence Representation}
Input sequence is transformed into feature vectors ($h_L$) using BERT's pre-trained language model, fine-tuned in our task. In spite of common practice to represent the sentence by the [CLS] vector in $h_L$ \cite{Alt2019}, we argue that not all words contribute equally to sentence representation. 

By encoding the underlying relation as a function of the examining entities and by giving attention to vectors related to this underlying relation, we can further reduce sentence noise and improve precision. Core modules constitute the: relation embedding, entities-wise attention, and relation attention. We examine them below.
\smallskip
\newline
\textbf{Relation Embedding}:
We formulate relation embeddings using the TransE model \cite{Bordes2013}. TransE model regards the embedding of the underlying relation $l$ as the distance (difference) between $h$ and $t$ embeddings ($l_i=t_i-h_i$), assuming that a relation $r$ holds between an entity pair ($h, t$). Then, we shape relation embedding for each sentence $i$ by applying a linear transformation on the head and tail entities vectors, activated through a Tanh layer to capture possible nonlinearities:
\begin{equation}
  l_{i} = Tanh(w_l(t_i - h_i) + b_l)
\end{equation}
, where $w_l$ is the underlying relation weight matrix and $b_l \in \Re^{d_t}$ is the bias vector.
We mark relation embedding as $l$ because it represents the possible underlying relation between the two entities and not the actual relationship $r$. Head $h_i$ and tail $t_i$ embeddings reflect only the entities' related tokens, which we capture through simple entities-wise attention, shown below.
\newline
\textbf{Entities-wise Attention}:
Head and tail embeddings participating in the relation embedding are created by summing over respective token vectors from BERT's last layer $h_L$. We capture these tokens through head- and tail-wise attention. Head-wise attention assigns the weight $\alpha_{it}^{h}$ to focus on head related tokens and tail-wise attention assigns the weight $\alpha_{it}^{t}$ to focus on tail related tokens. 
\begin{equation}
  \alpha_{it}^{h} =
    \begin{cases}
      1& \text{if $t= $ head in STP tokens} \\
      0&  \text{otherwise}
    \end{cases}       
\end{equation}

\begin{equation}
  \alpha_{it}^{t} =
    \begin{cases}
      1& \text{if $t= $ tail in STP tokens} \\
      0&  \text{otherwise}
    \end{cases}       
\end{equation}

Head $h_i$ and tail $t_i$ embeddings are then shaped as follows:

\noindent\begin{minipage}{.5\linewidth}
\begin{equation}
  h_i = \sum_{t=1}^{T}\alpha_{it}^{h}\cdot h_{it}
\end{equation}
\end{minipage}%
\begin{minipage}{.5\linewidth}
\begin{equation}
  t_i = \sum_{t=1}^{T}\alpha_{it}^{t}\cdot h_{it}
\end{equation}
\end{minipage}
\smallskip
\newline
\textbf{Relation Attention}:
Even though REDSandT is trained on STP that naturally preserves only relational tokens, we wanted to further reduce possible left noise on sentence-level. For this reason, we use a relation attention to emphasize on sentence tokens that are mostly related to the underlying relation $l_i$. We calculate relation attention $\alpha_r$ by comparing each sentence representation against the learned representation $l_i$ for each sentence $i$:
\begin{equation}
  \alpha_r=\frac{exp(s_i l_i)}{\sum_{j=1}^{n} exp(s_j l_i)}
\end{equation}

Then, we weight BERT' s last hidden layer $h_L \in \Re^{d_h}$ with relation embedding:
\begin{equation}
  h_L^{'} = \sum_{t=1}^{T}\alpha_{r}\cdot h_{it}
\end{equation}

Finally, \textbf{sentence representation} $s_i \in \Re^{d_h*2}$ is computed as the concatenation of the relation embedding $l_i$ and the sentence's weighted hidden representation $h_L^{'}$:
\begin{equation}
  s_i = \left[  l_i\; ; \; h_L^{'}  \right]
\end{equation}

Several other representation techniques were tested, with the presented method to outperform.

\subsection{BAG Encoder}
Bag encoding, i.e., aggregation of sentence representations in a bag, comes to reduce noise generated by the erroneously annotated relations accompanying DS.
Assuming that not all sentences contribute equally to the bag representation, we use selective attention \cite{Lin2016} to emphasize on sentences that better express the underlying relation.
\begin{equation}
    B = \sum_i \alpha_i s_i,
\end{equation}
As seen, selective attention represents bag as a weighted sum of the individual sentences. Attention $\alpha_i$ is calculated by comparing each sentence representation against a learned representation r:
\begin{equation}
    \alpha_i=\frac{exp(s_i r)}{\sum_{j=1}^{n} exp(s_j r)}
\end{equation}

Finally, bag representation $B$ is fed to a softmax classifier to obtain the probability distribution over the relations.
\begin{equation}
    p(r) = Softmax(W_r \cdot B + b_r), 
\end{equation}
where $W_r$ is the relation weight matrix and $b_r \in \Re^{d_r}$ is the bias vector. 

\subsection{Training}
REDSandT utilizes a transformer model, precisely BERT, which fine-tunes on our specific setup to capture the semantic features of relational sentences. Below, we present the overall process.

\subsubsection{Model Pre-training}
For our experiments, we use the pre-trained \textit{bert-base-cased} language model \cite{Devlin}, which consists of 12 layers, 12 attention heads, and 110M parameters, with each layer being a bidirectional Transformer encoder \cite{Vaswani2017}. The model is trained on cased English text of BooksCorpus and Wikipedia with a total of 800M and 2.5K words respectively. BERT is pre-trained using two unsupervised tasks: masked LM and next sentence prediction, with masked LM being its core novelty as it allows the previously impossible bidirectional training.

\subsubsection{Model Fine-tuning}
\begin{figure}[b]
    \centering
    \vspace{-4mm}
    \includegraphics[scale=0.49]{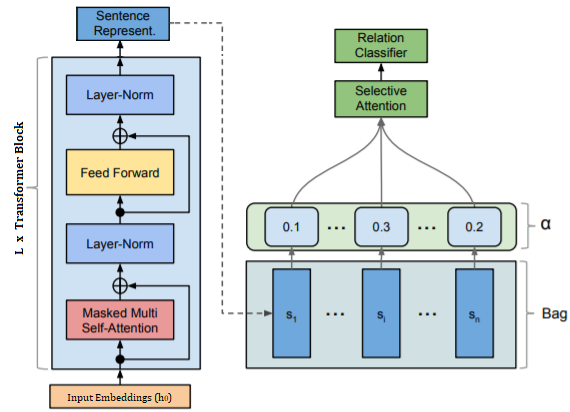}
    \caption{Transformer architecture (left) and training framework (right). Sentence representation $s_i$ is formed as shown in Figure \ref{fig:redsandt_input}.}
    \label{fig:redsandt_architecture}
    \vspace{-5mm}
\end{figure}

We initialize REDSandT model' s weights with the pre-trained BERT model, and we fine-tune its 4-last layers under the multi-instance learning setting presented in Figure \ref{fig:redsandt_architecture}, given the specific input shown in Figure \ref{fig:redsandt_input}. We end up fine-tuning only the last four layers after experimentation.

During fine-tuning, we optimize the following objective:
\begin{equation}
    L(D) = \sum_{i=1}^{|B|}log P(l_i|B_i ; \theta)
\end{equation}
, where for all entity pair bags $|B|$ in the dataset, we want to maximize the probability of correctly predicting the bag's relation given its sentences' representation and parameters.

\section{Experimental Setup}
\label{sect:experiments}

\subsection{Dataset}
We conduct experiments on the widely used benchmark dataset NYT-10 \cite{Riedel2010}, which was built by aligning triples in Freebase to the NYT corpus and contains 53 relations. There are 522,611 (172,448) sentences, 281,270 (96,678) entity pairs, and 18,252 (1,950) relation mentions in the train (test) set. We provide an enhanced dataset, \textit{NYT-10-enhanced}, including both STP and SDP versions of the input sentences as well as the head and tail entity types to facilitate future implementations.

\subsection{Hyper-parameter Settings}
\begingroup
\thickmuskip=0mu
\setlength{\thickmuskip}{0mu}
In our experiments we utilize \textit{bert-base-cased} model with hidden layer dimension $D_h=768$, while we fine-tune the model with \textit{max\_seq\_length} $D_t=64$. Regarding model's hyper-parameters, we manually tune them on the training set, based on AUC score. We select $batch\_size=32$ among $\{8, 16, 32\}$, $epochs=3$ among $\{3, 4\}$, BERT's fine-tuned $layers=4$ among all and last $\{2,4,8\}$, learning rate $lr=2e^{-5}$ among $\{2e^{-4}, 1e^{-5}, 2e^{-5}\}$, classifier dropout $p=0.4$ among $\{0.2, 0.4, 0.5\}$, and $weight\_decay=0.001$ among $\{0.01, 0.001\}$. Moreover, we fine-tune our model using the Adam optimization scheme \cite{Kingma2015} with $\beta_1 = 0.9, \beta_2 = 0.999$ and a cosine learning rate decay schedule with warm-up over 0.1\% of training updates. We minimize loss using cross entropy criterion weighted on dataset's classes to handle the unbalanced training set. Experiments conducted on a PC with 32.00 GB Ram, Intel i7-7800X CPU@ 3.5GHz and NVIDIA's GeForce GTX 1080 with 8GB. Training time takes $\sim$100min/epoch.
\endgroup

\subsection{State-of-the-art Models}
For evaluating REDSandT, we compare against following state-of-the-art models:
\newline
\textbf{Mintz} \cite{Mintz2009}: A multi-class logistic regression model under distant supervision setting.
\newline
\textbf{PCNN+ATT} \cite{Lin2016}: A CNN model with instance-level attention
\newline
\textbf{RESIDE} \cite{Vashishth2018}: A NN model that uses several side information (entity types\footnote{Compared to our 18 KB-independent entity types, authors use 38 Freebase-specific entity types.}, relational phrases) and employs Graph-CNN to capture syntactic information of instances. 
\newline
\textbf{DISTRE} \cite{Alt2019}: A transformer model, GPT fine-tuned for RE with an auxiliary objective under the distant supervision setting.


\section{Results}
\label{sec:results}

\subsection{Comparison with state-of-the-art Models}
\begin{figure}[H]
    \vspace{-2mm}
    \centering
    \includegraphics[scale=0.46]{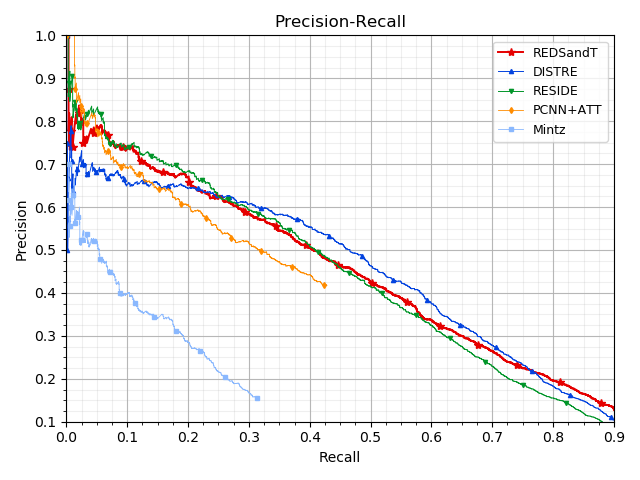}
    \setlength{\abovecaptionskip}{-5pt}
    \caption{Precision-Recall curves.}
    \label{fig:pr_curves}
    \vspace{-5mm}
\end{figure}
Figure \ref{fig:pr_curves} compares the precision-recall curves of REDSandT against state-of-the-art models. We observe that:
(1) The NN-based approaches outperform the probabilistic method (Mintz), showing human-designed features limitation against neural networks' automatically extracted features. (2) RESIDE, DISTRE, and REDSandT achieve better performance than PCNN+ATT, which even exhibiting the highest precision in the beginning soon follows an abrupt decline. This reveals the importance of both side-information (i.e., entity types and relation alias), and transfer knowledge. (3) RESIDE performs the best in low recalls and generally performs well, which we attribute to the multitude of side-information given. (4) Although DISTRE exhibits ~3.5\% greater precision in medium-level recalls, it presents 2-12\% lower precision in recall values $<$0.25 compared to RESIDE, and REDSandT. (5) Our model shows the more stable behavior, with a steady, downward trend, acting similar to RESIDE at the low and medium recalls and surpassing all baselines in the very high recall values. We believe the reason is that we use potential label information as an additional feature and as attention over the instance tokens. The learned label embeddings are of high quality since they carry common-knowledge from the pre-trained model fine-tuned on the specific dataset and task. Moreover, the chosen pre-trained model, BERT, considers bidirectionality while training, being thus able to efficiently capture head and tail interaction.

\begin{table}[t]
    \centering
    \begin{tabular}{@{} l  *4c @{}} 
    \toprule
     RE methods  & AUC & P@100 & P@300 & P@500\\
     \midrule
     Mintz  & 0.107 & 52.3 &  45.0 & 39.7\\
     PCNN+ATT & 0.341 & 73.0  & 67.3 & 63.6\\
     RESIDE & 0.415 & 81.8 & 74.3 & 69.7\\
     DISTRE & 0.422 & 68.0 & 65.3 & 65.0\\
     REDSandT & \textbf{0.424} & 78.0 & 73.0 & 67.6\\
    \bottomrule
    \end{tabular}
    \caption{AUC and P@N evaluation results. P@N represents precision calculated for the top N rated relation instances}
    \label{table:evalAgainstBaselines}
    \vspace{-9mm}
\end{table}
Table \ref{table:evalAgainstBaselines}, which presents AUC and precision at various points in the P-R curve, reveals our model's precision performance to be between that of RESIDE and DISTRE while preserving the state-of-the-art AUC. Precisely, REDSandT' s precision does not exceed RESIDE', even though it is close enough, which suggests that additional side-information would improve our model. Meanwhile, REDSandT surpasses DISTRE' s precision, which we attribute to our selected pre-trained model that efficiently captures label embeddings. Consequently, our model is more consistent to the various points of the P-R curve.

Table \ref{table:relDistributionTop300} shows the distribution over relation types for the top 300 predictions of REDSandT and baseline models. REDSandT encompasses 10 distinct relation types, two of which (\textit{place\_founded}, \textit{/geographic\_distribution}) are not recognized by none of rest models. PCNN+ATT predictions are highly biased towards a set of only four relation types, while RESIDE captures three additional types. DISTRE and REDSandT manage to recognize more types than all models, emphasizing the contribution of transfer knowledge. Moreover, REDSandT correctly not recognizes \textit{/location/country/capital} relation that DISTRE does, as their authors found most errors to arise from the specific predicted relation in manual evaluation. 
Meanwhile, we highlight REDSandT' s effectiveness in recognizing relations in the long-tail. Particularly, our model captures, \textit{founders} (1.47\%), \textit{neighborhood\_of} (1.06\%), \textit{person/children} (0.47\%), and \textit{sports\_team/location} (0.16\%) relations. Relations are listed in descending order regarding population in test set with respective percentage referenced in parentheses.

\begin{table}[t]
\begin{tabular}{@{}l*4c@{}} 
\toprule
 Relation  & red & dis & res & pcnn \\
 \midrule
 /location/contains   & 176 & 168 & 182& 214\\
 /person/company      & 38  & 31  & 26 & 19\\
 /person/nationality  & 26  & 32  & 65 & 59\\
 /admin\_div/country  & 25  & 13  & 12 & 6\\
 /neighborhood\_of    & 22  & 10  & 3  & 2\\
 /person/children     & 5   &  -  & 6  & -\\
 /team/location       & 4   & 2   &  - & -\\
 /founders            & 2   &  2  & 6  & -\\
 /place\_founded      & 1   &  -  & -  & -\\
 /geo\_distribution   & 1   &  -  & -  & -\\
 /country/capital     & -   & 17  & -  & -\\
 /person/place\_lived & -   & 22  & -  & -\\
\bottomrule
\end{tabular}
\caption{Relation Distribution over the top 300 predictions for PCNN+ATT(pcnn), RESIDE(res), DISTRE(dis) and REDSandT(red) models}
\label{table:relDistributionTop300}
\vspace{-2mm}
\end{table}

\subsection{Ablation Studies}
\label{sec:ablation_studies}
To assess the effectiveness of the different modules of REDSandT, we create four ablation models:
\newline
\textbf{REDSandT w/o ET}: Removes entity types from input sentence representation.
\newline
\textbf{REDSandT w/o $r_{ht}$}: Removes relation embedding and relation attention. We represent sentence using the [CLS] token of BERT's last hidden layer $h_L$ . 
\newline
\textbf{REDSandT w/o $a_r$}: Removes relation attention on instance tokens.
\newline
\textbf{REDSandT w. SDP}: Replaces STP with SDP \cite{Xu2015} in sentence encoding.

\begin{table}[t!]
\begin{tabular}{l c c c c}
\toprule
\multirow{2}{*}{\thead{Metrics}} &
\multirow{2}{*}{\thead{AUC}} &
\multicolumn{3}{c}{\thead{P@N(\%)}} \\
\cmidrule(lr){3-5}
&& 100 & 200 & 300 \\
\midrule
\small{REDSandT w/o} $r_{ht}$ & 0.404 & 80.0 & 72.0 & 67.7\\
\small{REDSandT w/o} ET       & 0.415 & 78.0 & 74.0 & 71.3\\
\small{REDSandT w.} SDP       & 0.418 & 75.0 & 71.0 & 69.7\\
\small{REDSandT w/o} $a_r$    & 0.422 & 75.0 & 76.0 & 71.0\\
\small{REDSandT}              & 0.424 & 78.0 & 75.0 & 73.0 \\
\bottomrule
\hline
\end{tabular}
\caption{Evaluation results AUC and P@N of variant models on NYT-10 dataset.}
\label{table:evalInAblationCases}
\vspace{-5mm}
\end{table}

As shown in Table \ref{table:evalInAblationCases}, all modules contribute to final model' s effectiveness. Greatest impact comes from relation embeddings with their removal resulting in the highest AUC (2 units) and P@300 (5.3\%) drop. Meanwhile, P@100 goes up to 80\% with inspection of top 300 predictions revealing a focus on 5 relation types only, with \textit{/location/contains} to make up the 79\% of these. Simple integration of entity types in input representation is the next most important feature that boosts our model. Next, ``REDSandT w. SDP", shows STP's superiority, while a manual inspection in the model's top 300 predictions prove SDP's weakness to recognize relations in the long tail, with focus given on \textit{/person/nationality} relation. Finally, removing the relation attention over instance tokens exhibits the least effect in AUC (0.002) and precision ($\sim$2\%). Meanwhile, we notice that model focuses solely on 8 relation types in the top 300 predictions.

\subsection{Case study: Effect of relation attention}
\begin{figure}[H]
    \vspace{-4mm}
    \centering
    \includegraphics[scale=0.38]{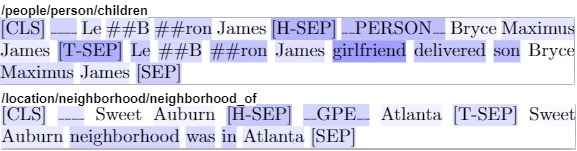}
    \setlength{\abovecaptionskip}{-8pt}
    \caption{Relation attention weights for \textit{children} (top) and \textit{neighborhood\_of} (bottom) long-tail relations.}
    \label{fig:rel_attention_weights}
    \vspace{-3mm}
\end{figure}
\raggedbottom

Figure \ref{fig:rel_attention_weights} shows a visualization 
 of the relation attention weights, highlighting the different parts of the sentence that drive relation extraction, for two long-tail relations. In both cases, we see that the special tokens preserve important information, while also the entity type is given more weight than the entity itself. Moreover, we see which tokens affect more the relation. Tokens ``girlfriend", ``son", and the repetition of name ``James" are predictive of the ``\textit{children}" relation, while tokens ``neighborhood", ``was", ``in", along with a GPE entity type show a probable ``\textit{neighborhood\_of}" relation.

\section{Related Work}
\label{sect:relwork}
Our work is related to distant supervision, neural relation extraction (mainly pre-trained LMs), sub-tree parse of input, label embedding, and entity type side information.
\newline
\textbf{Distant Supervision}:
DS plays a key role in RE, as it satisfies its need for extensive training data, easily and inexpensively. The use of DS \cite{Craven1999, Snow2005} to generate large training data for RE was proposed by \cite{Mintz2009}, who assumed that all sentences that include an entity pair, which exhibits a relationship in a KB, express the same relation. However, this assumption comes with noisy labels, especially when the KB is not directly related to the domain at hand. Multi-instance learning methods were proposed to alleviate the issue, by conducting relation classification at the bag level, with a bag including instances that mention the same entity pair \cite{Riedel2010, Hoffmann_2011}. 
\newline
\textbf{Neural Relation Extraction}:
While the performance of the above approaches heavily relies on handcrafted features (POS tags, named entity tags, morphological features, etc.), the advent of neural networks in RE set the focus on model architecture. 
\newcite{Zeng2014a} propose a CNN-based method to automatically capture the semantics of sentences, while PCNN \cite{Zeng2015} became the common architecture to embed sentences. PCNN is used in several approaches that handle DS noisy patterns, such as intra-bag attention \cite{Lin2016}, inter-bag attention \cite{Ye2019}, soft labeling \cite{Liu2017a, Wang2018} and adversarial training \cite{Wu2018,qin2018dsgan}. Moreover, Graph-CNNs proved an effective way to encode syntactic information from text \cite{Vashishth2018}.

The latest development of pre-trained LMs relying on transformer architecture \cite{Vaswani2017} has shown to capture semantic and syntactic features better \cite{Radford2018}. \citet{Howard2018} found that they significantly improve text classification performance, prevent overfitting, and increase sample efficiency. \citet{Shi2019} fine-tuned BERT \cite{Devlin} on the TACRED dataset showing that simple NNs built on top of BERT improve performance. Meanwhile, \citet{Alt2019} extended GPT \cite{Radford2018} to the DS setting by incorporating a multi-instance training mechanism, proving that pre-trained LMs provide a stronger signal for DS than specific linguistic and side-information features \cite{Vashishth2018}. 
\newline
\textbf{Side information}:
Apart from model architecture, several methods propose additional information to further reduce noise. \citet{Vashishth2018} use relation phrases and incorporate Freebase entity types achieving state-of-the-art precision at higher recall values, while \cite{Ji2017, Hu2019} use entity descriptors to enhance entity and label embeddings, respectively. 
\newline
\textbf{Sub-Parses of Input}:
\citet{Xu2015} showed the importance of the shortest-dependency path (SDP) in reducing irrelevant to RE words. \citet{Liu2018} further reduce the noise within sentences by preserving the sub-path of the sentence that connects the two entities with their least common ancestor's parent (STP). In contrast with \cite{Liu2018}, who shape the final STP sequence by re-assigning the participating tokens into their original sequence order, we preserve the tokens' order within the STP to maintain the emerged grammar information.
\newline
\textbf{Label Embedding}:
Label embeddings aim to embed labels in the same space with word vectors. The idea comes from computer vision, with \cite{Wang2018} to introduce them in text classification and \cite{Hu2019} to use them as attention-mechanism over relational tokens in distantly-supervised RE. We make use of the TransE \cite{Bordes2013} model to shape label embeddings as the entities' distance in BERT's vector space, and we show that their use both as a feature and as attention over sentences significantly improves RE.



\section{Conclusion}
\label{sect:conclusion}
We presented a novel transformer-based relation extraction model for distant supervision. REDSandT manages to acquire high-informative instance and label embeddings and is efficient at handling the noisy labeling problem of DS. REDSandT captures high-informative embeddings for RE by fine-tuning BERT on a RE-specific structured input that focuses solely on relational arguments, including the sub-tree connecting the entities along with entities' types. Then, it utilizes these vectors to encode label embeddings, which are also used as attention mechanism over instances to reduce the effect of less-informative tokens. Finally, relation extraction occurs at bag-level by concatenating label and weighted instance embeddings. 
Extensive experiments on the NYT-10 dataset illustrate REDSandT's effectiveness over existing baselines in current literature. Precisely, REDSandT manages to recognize relations that other methods fail to detect, including relations in the long-tail.
Future work includes an investigation of whether additional information, such as entity descriptors, influence REDSandT's performance and to what extent, while also whether the special token embeddings can act as global embeddings for RE.

\bibliographystyle{acl_natbib}

\end{document}